\documentclass[conference]{IEEEtran}
\IEEEoverridecommandlockouts
\usepackage{cite}
\usepackage{amsmath,amssymb,amsfonts}
\usepackage{algorithmic}
\usepackage{graphicx}
\usepackage{textcomp}
\usepackage{xcolor}
\usepackage{colortbl}
\usepackage{booktabs}
\usepackage{multirow}
\usepackage{tabularx} 
\def\BibTeX{{\rm B\kern-.05em{\sc i\kern-.025em b}\kern-.08em
    T\kern-.1667em\lower.7ex\hbox{E}\kern-.125emX}}
\begin{document}

\title{Efficient Oriented Object Detection with Enhanced Small Object Recognition in Aerial Images}

\author{\IEEEauthorblockN{Anonymous Authors}}
\author{\IEEEauthorblockN{1\textsuperscript{st} Zhifei Shi}
\IEEEauthorblockA{\textit{School of Artificial Intelligence}\\
\textit{Beijing Normal University}\\
Beijing, China \\
202321081018@mail.bnu.edu.cn}
\and
\IEEEauthorblockN{2\textsuperscript{nd} Zongyao	Yin}
\IEEEauthorblockA{\textit{School of Artificial Intelligence}\\
\textit{Beijing Normal University}\\
Beijing, China \\
202231081006@mail.bnu.edu.cn}
\and
\IEEEauthorblockN{3\textsuperscript{rd} Sheng Chang}
\IEEEauthorblockA{\textit{School of Artificial Intelligence}\\
\textit{Beijing Normal University}\\
Beijing, China \\
changsheng@mail.bnu.edu.cn}
\and
\IEEEauthorblockN{4\textsuperscript{th} Yi Xiao}
\IEEEauthorblockA{\textit{School of Artificial Intelligence}\\
\textit{Beijing Normal University}\\
Beijing, China \\
2182772477@qq.com}
\and
\IEEEauthorblockN{7\textsuperscript{th} XianChuan Yu}
\IEEEauthorblockA{\textit{School of Artificial Intelligence}\\
\textit{Beijing Normal University}\\
Beijing, China \\
yuxianchuan@163.com}
}

\maketitle

\begin{abstract}
Achieving a balance between computational efficiency and detection accuracy in the realm of rotated bounding box object detection within aerial imagery is a significant challenge. While prior research has aimed at creating lightweight models that enhance computational performance and feature extraction, there remains a gap in the performance of these networks when it comes to the detection of small and multi-scale objects in remote sensing (RS) imagery. To address these challenges, we present a novel enhancement to the YOLOv8 model, tailored for oriented object detection tasks and optimized for environments with limited computational resources. Our model features a wavelet transform-based C2f module for capturing associative features and an Adaptive Scale Feature Pyramid (ASFP) module that leverages P2 layer details. Additionally, the incorporation of GhostDynamicConv significantly contributes to the model's lightweight nature, ensuring high efficiency in aerial imagery analysis. Featuring a parameter count of 21.6M, our approach provides a more efficient architectural design than DecoupleNet, which has 23.3M parameters, all while maintaining detection accuracy. On the DOTAv1.0 dataset, our model demonstrates a mean Average Precision (mAP) that is competitive with leading methods such as DecoupleNet. The model's efficiency, combined with its reduced parameter count, makes it a strong candidate for aerial object detection, particularly in resource-constrained environments.
\end{abstract}

\begin{IEEEkeywords}
Rotated Detection, Small Object Recognition, Lightweight Model, Aerial Image Processing
\end{IEEEkeywords}

\section{INTRODUCTION}
Rotated bounding box object detection is a crucial technology in the field of computer vision, especially in remote sensing image analysis, where it holds significant application value and practical demand\cite{obb1,OBB4}. With the continuous advancement of deep learning technology, rotated bounding box detection has shown its unique advantages in handling objects with rotational and inclined characteristics\cite{obb,OBB5}, such as reducing background information within the bounding box and better distinguishing densely arranged objects, thus reducing false detections and missed detections\cite{obb2}.

Despite advancements in feature extraction and lightweight design, existing methods continue to face challenges such as varying object scales, dense arrangements, and small object detection. Techniques like Spatial Pyramid Pooling (SPP) \cite{SPP} and its optimized variant SimSPPF \cite{simSPPF} improve adaptability to different input sizes, enhancing the model's ability to handle complex scenarios. Feature Pyramid Networks (FPN) \cite{FPN} and Path Aggregation Networks (PANet) \cite{PAFPN} further enhance multi-scale feature extraction by merging feature maps from various levels, although they may lose some low-level details. Lightweight design approaches, such as the GhostNet series \cite{han2020ghostnet,tang2022ghostnetv2}, utilize dynamic convolution with multiple expert networks to enhance parameter efficiency and feature representation capability. Expanding on this, lightweight networks tailored for remote sensing (RS) visual tasks, exemplified by MSDWNet\cite{msdw} and LO-Det\cite{lodet}, are engineered to augment both computational efficiency and detection accuracy, especially for multiscale objects. Despite their commendable efforts, these networks often grapple with the complexities of feature extraction, a critical aspect that can significantly hinder performance. Wavelet transform methods \cite{yoo2019photorealistic,14} offer advantages in reducing computational complexity while maintaining or improving detection performance. Efforts to improve small object detection by adding a P2 layer \cite{P2,P21,P22} have increased computational complexity, highlighting an unresolved tension between detection accuracy and efficiency. The DecoupleNet\cite{lu2024decouplenet}, a recent approach in this domain, struggles with the extraction of detailed features effectively, a limitation that hinders its performance in complex aerial scenes. To address these limitations, this study introduces the Adaptive Scale Feature Pyramid (ASFP) and the C2f with Next-Gen Convolution feature extraction structures, aimed at improving small object detection performance while maintaining a low parameter count.

Synthesizing our contributions, this paper introduces three transformative methodologies in aerial object detection. We deliver: 1) A refined C2f module that sharpens feature extraction; 2) The pioneering Adaptive Scale Feature Pyramid (ASFP) for heightened acuity in small object identification; and 3) A lightweight framework that significantly prunes parameter volume, thereby streamlining computational efficiency. These innovations enhance detection capabilities, align with real-world needs, and lay a solid foundation for the detailed methodology and empirical findings presented later.

\section{METHODOLOGY}
\begin{figure*}[htbp]
  \centering
  \includegraphics[width=0.9\linewidth]{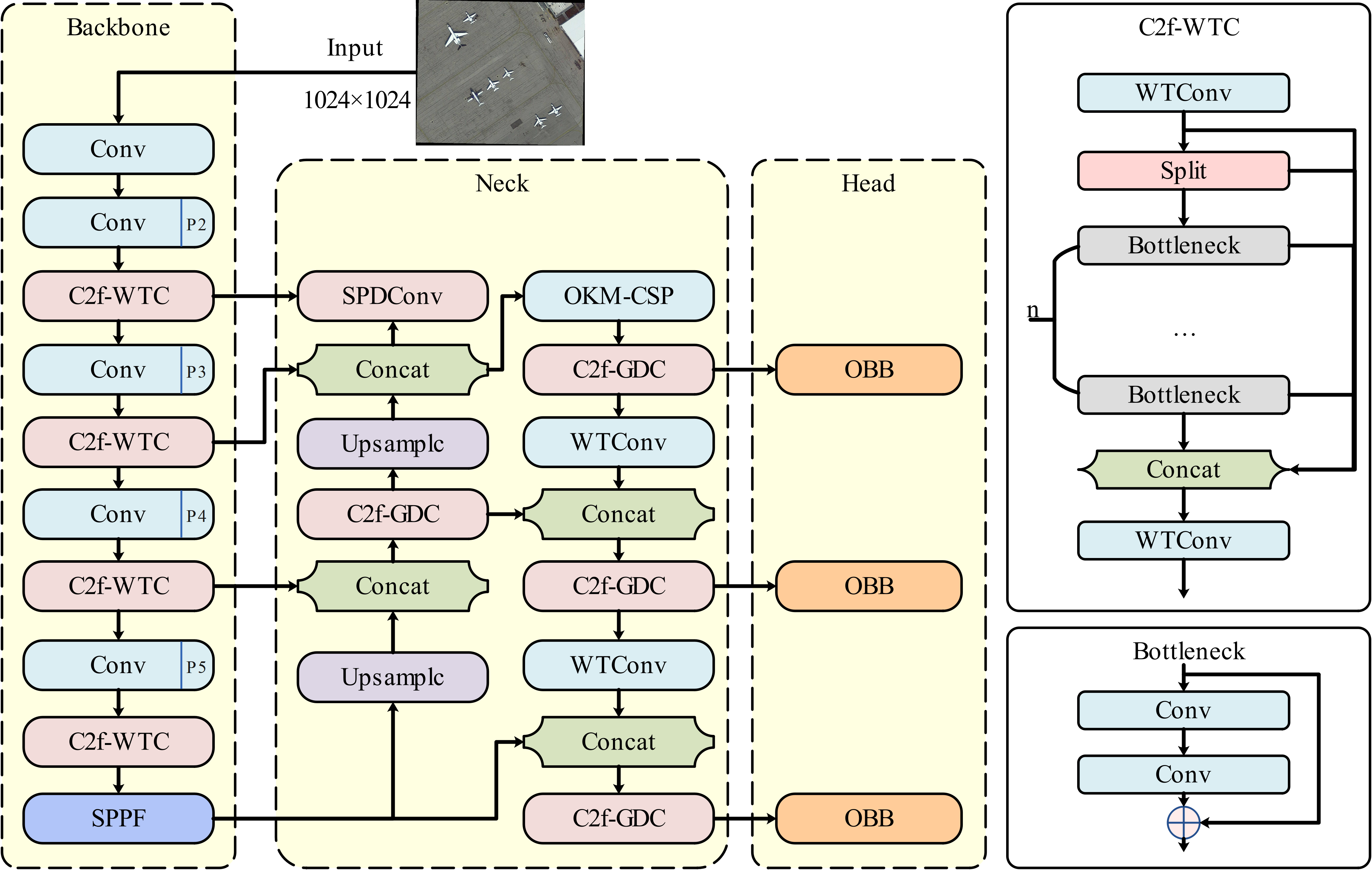}
  \caption{Overview of our model}
  \label{fig:example}
\end{figure*}

To achieve the goal of enhancing the model's ability to detect small objects while maintaining a low parameter count, we propose our network structure, as shown in Figure \ref{fig:example}.
\subsection{Small Object Enhance Pyramid}
To address the poor performance of traditional YOLO-based improvements in detecting low-resolution images and small objects, we designed the Adaptive Scale Feature Pyramid (ASFP) structure in the feature extraction stage of YOLOv8. The ASFP structure is designed to leverage the strengths of both P2 and P3 layers in the backbone network. The P2 layer, known for its higher resolution, provides rich feature details that are crucial for small object detection. Conversely, the P3 layer offers a balance between resolution and contextual information, making it suitable for capturing medium-scale features.

To circumvent the computational burden of introducing a P2 detection layer, we harnessed SPDConv\cite{SPDConv} to distill small object-enriched features from the P2 layer of our backbone. These features are merged with those from the P3 detection layer, providing a comprehensive feature set to the OKM-CSP module, thereby enhancing computational efficiency while retaining the sensitivity to small objects.

We identified the original CSP structure's efficacy in reducing computational load and enhancing feature diversity through feature map splitting and merging, but noted its shortcomings in small object detection. To address this, inspired by the Omni-Kernel Module (OKM) \cite{cui2024omni}, we introduced the OKM-CSP module,incorporating specialized branches for small object feature extraction. Utilizing the parameter $e$ to regulate channel allocation, the input feature map $X$ is bifurcated into the okbranch and identity, with respective channel ratios of $e$ and $(1-e)$. The okbranch output is further channeled into three parallel branches—Global, Large, and Local—to refine small object feature extraction.

{\bf Global Branch}: This branch focuses on extracting features with large receptive fields to capture global contextual information. Specifically, it uses Dual-domain Channel Attention Module (DCAM) and a Frequency-based Spatial Attention Module (FSAM) to expand the receptive field without increasing the number of parameters significantly.

{\bf Large Branch}: This branch is designed to capture medium-scale features. It employs standard convolutional layers with a moderate receptive field size, ensuring a balance between global context and local details.

{\bf Local Branch}: This branch focuses on extracting fine-grained features crucial for small object detection. It uses standard convolutional layers with small kernel sizes to capture detailed local information.

\begin{figure}[htbp]
  \centering
  \includegraphics[width=0.9\linewidth]{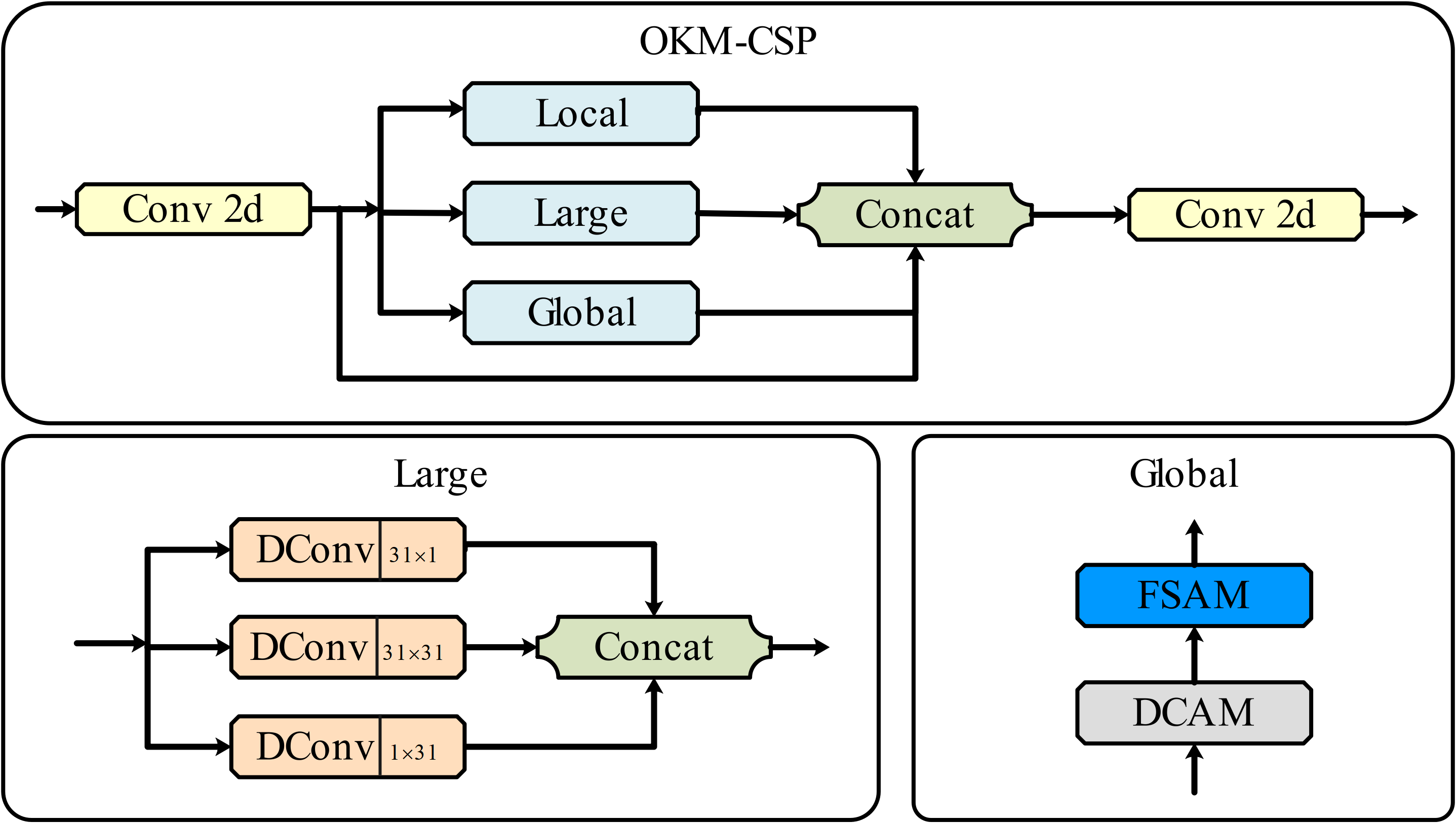}
  \caption{Overview of OKM-CSP}
  \label{fig2}
\end{figure}

The outputs of the three branches are then concatenated and merged with the identity branch, through a convolutional layer, and finally obtain the output feature map, as illustrated in Figure \ref{fig2}.

Assuming the number of channels in the input feature map is $C$, the number of channels in the okbranch is $Ce$, and the number of channels in the identity branch is $C(1-e)$. The input feature map  $x$ is divided into two parts:\begin{equation}x^{'}=\frac{e}{1+e}x,\end{equation} and \begin{equation}x^{''}=\frac{1}{1+e}x.\end{equation} Where $x^{\prime}$ is the feature map input to the OmniKernel module, and $x^{\prime\prime}$ is directly connected to the next layer. The parameter $e$ controls the split ratio. For better performance, we choose $e=0.25$.

Next, the OmniKernel module receives $x^{\prime}$ and outputs the feature map $y^{\prime}=\text{OminKernel}(x^{\prime})$. Finally, the output $\mathrm{y}^{\prime}$ from the OmniKernel module is concatenated with $x^{\prime\prime}$ and passed through the convolutional layer for feature fusion, resulting in the final output feature map \begin{equation}y=\text{Conv}2d\left(\text{Concat}(y^{\prime},x^{\prime\prime})\right).\end{equation}

By introducing the OKM-CSP module, we can enhance the model's ability to extract features at multiple scales and improving the detection accuracy of small objects.
\subsection{C2f with Next-Gen Convolution}
The Cross-stage Partial Feature (C2f) module is a crucial component of the YOLOv8 model, designed to enhance the feature extraction process. This module facilitates partial connections across different stages of the network, enabling a more integrated and comprehensive representation of features. Such an approach is particularly advantageous for object detection, as it allows for a more detailed and nuanced assimilation of features from various levels within the network.

Nevertheless, the C2f module has notable limitations, especially in detecting small objects, which warrant further investigation. The localized nature of the module’s convolution operations may prove inadequate for extracting features from complex images that require a broader receptive field. Furthermore, the parameter count of the C2f module increases quadratically with the size of the convolution kernel, which raises concerns regarding over-parameterization. This issue can adversely affect the computational efficiency and scalability of the model, underscoring the necessity for a more refined approach to feature extraction that strikes a balance between detailed feature representation and the constraints of computational resources.
\subsubsection{C2f-Wavelet Transform Convolution (C2f-WTC)}
The C2f module in YOLOv8 is pivotal for object detection but encounters limitations in feature extraction due to localized convolution operations, particularly with complex images. To address these challenges, we introduce the C2f-Wavelet Transform Convolution (C2f-WTC), which leverages wavelet transforms to enhance receptive fields while minimizing parameter growth.

Utilizing the Haar wavelet for its simplicity, we decompose an image $X$ of dimensions $N_{w}\times N_{h}$ into four sub-bands: $X_{LL}$ (low-frequency), $X_{LH}$, $X_{HL}$, and $X_{HH}$ (high-frequency). The transformation process focuses on extracting relevant features across varying scales. The Haar wavelet transform can be expressed as follows:
\begin{equation}X_{LL}^{(i)},X_{LH}^{(i)},X_{HL}^{(i)},X_{HH}^{(i)}=\mathrm{WT}\left(X_{LL}^{(i-1)}\right),\end{equation}
where $i$ denotes the current decomposition level. By recursively applying the Haar WT to $X_{LL}$, we create a cascade of wavelet transforms that effectively capture both low and high-frequency components of the input image, facilitating detailed feature extraction.

In the WTConv approach, the input $X$ is transformed using wavelet techniques before applying convolution with small kernels $W$ across each frequency sub-band. This process is represented as:
\begin{equation}Y=\mathrm{IWT}\left(\mathrm{Conv}(W,\mathrm{WT}(X))\right),\end{equation}
where $\mathrm{IWT}$ denotes the inverse wavelet transform, which recombines the transformed frequency components back into the spatial domain. This technique allows the model to integrate multi-scale features effectively, enhancing its sensitivity to both fine and coarse details.

Incorporated within C2f-WTC, the Bottleneck structure serves a dual purpose: it streamlines the feature processing pipeline and curbs computational expenditure.Essentially, the Bottleneck acts as a filter, narrowing down the feature maps to their most salient aspects through an initial $1\times1$ convolution. This condensed representation is then subjected to a $3\times3$ convolution, which further refines the features before they are expanded back to their original dimensionality. The strategic reduction and subsequent expansion of feature maps enable the WTC module to maintain a high level of feature discrimination with reduced computational load, which is essential for the accurate detection of small objects in aerial imagery.

The key advantage of C2f-WTC is its logarithmic growth of parameters relative to the receptive field size, contrasting with traditional methods that exhibit quadratic growth. This design refines detection precision and parameter thriftiness, adeptly handling complex imagery with an economical computational footprint, thus elevating the model's oriented object detection capabilities.

\subsubsection{C2f-Ghost Dynamic Convolution (C2f-GDC)}
Given the computational and complexity constraints of the C2f module, we integrated the GhostModule\cite{han2024parameternet} to augment the model's capacity with minimal additional computation. The GhostModule employs dynamic convolution with expert networks for feature processing, achieving enhanced expressiveness and efficiency through dynamic coefficient-based weighted summation.

Let $X$ be the input feature map, $W_{i}$ be the weight of the $i$-th expert, and $\alpha _{i}$ be the corresponding dynamic coefficient. The dynamic convolution can be represented as:
\begin{equation}Y_\text{dynamic}=\sum_{i=1}^M\alpha_i\odot\mathrm{Conv}(W_i,X),\end{equation}
where $M$ is the number of experts, and the dynamic coefficients $a_{i}$ are generated as $\alpha=\mathrm{softmax}\Big(\mathrm{MLP}\big(\mathrm{Pool}(X)\big)\Big)$, where $\operatorname{Pool}(X)$ is the pooling operation on the input X to reduce its spatial dimensions, and $MLP$ is a small two-layer feedforward network that converts the pooled vector into the coefficient vector $\alpha$, ensuring the sum of all coefficients equals 1 through the softmax function.

Combining the C2f module and GhostModule, we propose the C2f-GDC structure. In this structure, the convolution operations in each Bottleneck block are replaced by dynamic convolutions. Let $Y_{\mathrm{dynamic}1}$ and $Y_{\mathrm{dynamic}2}$ be the outputs of the first and second dynamic convolutions, respectively, the C2f-GDC module can be represented as:
\begin{equation}Y_{\mathrm{C2f-GDC}}=\mathrm{Conv}(Y_{\mathrm{dynamicl}}+X)+X,\end{equation}
where $X$ is the input of the residual connection, and $\mathrm{Conv}$ is the second convolution layer.

In C2f-GDC, each dynamic convolution introduces $M$ additional convolution kernels $D_{i}$, but these kernels share the same input and output dimensions, so the increase in parameters is proportional to $M$ and independent of the kernel size. Moreover, since the computational cost of dynamic coefficients is relatively low, the overall increase in computational load (FLOPs) is limited.

\section{EXPERIMENTS}
\subsection{Dataset Details}
In the context of oriented object detection within aerial imagery, the DOTAv1.0\cite{xia2018dota}, DIOR-R\cite{dioraopg}, and UCAS-AOD\cite{ucas} datasets have been selected for their comprehensive representation of small target detection challenges and their widespread use in the academic community.

\subsubsection{DOTAv1.0 Dataset}This dataset comprises 2,806 aerial images with 188,282 annotated instances across 15 categories. It is segmented into training and testing sets, with approximately 21,046 patches for training and 10,833 for testing, each patch being 1024x1024 pixels with a 200-pixel overlap. This setup is pivotal for assessing our model's capability to detect small objects amidst the complexity of aerial scenes.

\subsubsection{DIOR-R Dataset}Featuring 23,463 images, the DIOR-R dataset is instrumental for evaluating small object detection. It contains a variety of object sizes and backgrounds, providing a challenging environment that mirrors real-world conditions.

\subsubsection{UCAS-AOD Dataset}With 1,510 images and 7,482 annotated plane objects, this dataset is particularly valuable for its focus on aircraft and vehicle detection. It offers a uniform distribution of object orientations, which is critical for developing models that must be robust to various viewing angles.

The selection of these datasets is strategic, offering a diverse range of object sizes and orientations. This diversity is essential for testing the mettle of our detection model, ensuring it performs reliably even when tasked with identifying small targets in expansive aerial views.

\subsection{Experimental Setup}

In our experimental setup, we meticulously prepared the datasets and training parameters to ensure a rigorous evaluation of our proposed method. We conducted the backbone pretraining on the ImageNet-1k\cite{deng2009imagenet} for 300 epochs. For the DOTAv1.0\cite{xia2018dota} dataset, which serves as a benchmark for oriented object detection in aerial images, we set the input image size to $1024 \times 1024$ pixels and trained the models for 100 epochs. This resolution is standard for this dataset, allowing for a comprehensive assessment of our model's detection capabilities. For the DIOR-R\cite{dioraopg} dataset, we adjusted the input image size to $800 \times 800$ pixels and trained the models for 100 epochs. The UCAS-AOD\cite{ucas} dataset was processed by resizing the images to $640 \times 640$ pixels, and models were trained for 100 epochs. The UCAS-AOD dataset was divided into training, validation, and test sets according to the DOTA's standard split ratio of 5:2:3. Specifically, the training set contains 755 images, the validation set has 302 images, and the test set comprises 452 images.

For training the comparative methods, we adhered to the hyperparameters as detailed in their original publications, ensuring a fair and consistent comparison across all evaluated models.

\subsection{Experimental Results}
This section delineates the outcomes of our methodological exploration, offering a comprehensive evaluation of our proposed model's performance on the DOTAv1.0\cite{xia2018dota}, DIOR-R\cite{dioraopg}, and UCAS-AOD\cite{ucas} datasets. Our analysis focuses on the model's oriented object detection capabilities, particularly its proficiency in recognizing small-scale objects within aerial imagery. The subsequent subsections break down the performance metrics, comparative analyses, and ablation studies, providing a thorough examination of our model against existing benchmarks.

Table \ref{dota} provides a detailed evaluation of our model's performance on the DOTAv1.0 dataset, highlighting its capabilities in oriented object detection within aerial scenes. Our YOLOv8s model demonstrates a remarkable mean Average Precision (mAP) of 78.32\%, significantly surpassing $\text{S}^2\text{A}\text{-Net}$\cite{S2ANET}’s performance at 76.1\%, and demonstrating its robustness across various object categories.

In the category of small vehicles (SV), our model excels with an AP of 70.29\%, substantially surpassing $\text{S}^2\text{A}\text{-Net}$\cite{S2ANET}'s 65.03\%. Similarly, for storage tanks (ST), our model shows a marked improvement with an AP of 75.60\% compared to $\text{S}^2\text{A}\text{-Net}$\cite{S2ANET}'s 65.53\%. These enhancements are crucial for applications requiring precise detection of small objects in expansive aerial views.


\begin{table*}[!ht]
    \centering
    \footnotesize 
    \setlength{\tabcolsep}{2.5pt} 
    \caption{Experimental results of the baseline network on the DOTAv1.0 dataset. LV: LARGE VEHICLE. SP: SWIMMING POOL. HC: HELICOPTER. BR: BRIDGE. PL: PLANE. SH: SHIP. SBF: SOCCER-BALL FIELD. BC: BASKETBALL COURT. GTF: GROUND TRACK FIELD. SV: SMALL VEHICLE. BD: BASEBALL DIAMOND. TC: TENNIS COURT. RA: ROUNDABOUT. ST: STORAGE TANK. HA: HARBOR}
    \begin{tabular}{ccc|ccccccccccccccc}
    \hline
       \multirow{2}{*}{\bf Methods} & \multirow{2}{*}{\bf Backbone} & \multirow{2}{*}{\bf mAP(\%)$\uparrow$} & \multicolumn{15}{c}{\bf AP per category(\%)$\uparrow$} \\ \cline{4-18}
        ~ & ~ & ~ & PL & BD & BR & GTF & SV & LV & SH & TC & BC & ST & SBF & RA & HA & SP & HC \\ \hline
        PIoU\cite{piou_2020_chen} & DAL-34\cite{centernet} & 60.50 & 80.90 & 69.70 & 24.10 & 60.20 & 38.30 & 64.40 & 64.80 & \bf 90.90 & 77.20 & 70.40 & 46.50 & 37.10 & 57.10 & 61.90 & 64.00 \\ 
        $\text{O}^2\text{-Dnet}$\cite{o2dnet} & H-104\cite{centernet} & 71.04 & 89.31 & 82.14 & 47.33 & 61.21 & 71.32 & 74.03 & 78.62 & 90.76 & 82.23 & 81.36 & 60.93 & 60.17 & 58.21 & 66.98 & 61.93  \\
        Rn-psc\cite{rnpsc} & ResNet50\cite{resnet} & 71.09 & 89.32 & 82.29 & 37.92 & 71.52 & 78.40 & 66.33 & 78.01 & 90.89 & 84.21 & 80.63 & 60.22 & 64.73 & 59.69 & 68.37 & 53.85  \\
        $\text{P-RSDet}$\cite{prsdet} & ResNet101\cite{resnet} & 72.30 & 88.58 & 77.83 & 50.44 & 69.29 & 71.10 & 75.79 & 78.66 & 90.88 & 80.10 & 81.71 & 57.92 & 63.03 & 66.30 & 69.77 & 63.13  \\
        $\text{BBAVec}$\cite{bbavec_2021_yi} & ResNet101\cite{resnet} & 72.32 & 88.35 & 79.96 & 50.69 & 62.18 & 78.43 & 78.98 & 87.94 & 90.85 & 83.58 & 84.35 & 54.13 & 60.24 & 65.22 & 64.28 & 55.70 \\ 
        SCRDet\cite{yang2019scrdet} & ResNet50\cite{resnet} & 72.61 & 89.98 & 80.65 & 52.09 & 68.36 & 68.36 & 60.32 & 72.41 & 90.85 & 87.94 & \bf 86.86 & 65.02 & 66.68 & 66.25 & 68.24 & 65.21 \\ 
        RoI Trans.\cite{ROITRANS} & ResNet50\cite{resnet} & 74.05 & 89.01 & 77.48 & 51.64 & 72.07 & 74.43 & 77.55 & 87.76 & 90.81 & 79.71 & 85.27 & 58.36 & 64.11 & 76.50 & 71.99 & 54.06 \\ 
        G.V.\cite{xu2020gliding} & ResNet50\cite{resnet} & 75.02 & 89.64 & \bf 85.00 & 52.26 & \bf 77.34 & 73.01 & 73.14 & 86.82 & 90.74 & 79.02 & 86.81 & 59.55 & 70.91 & 72.94 & 70.86 & 57.32 \\ 
        AOPG\cite{dioraopg} & ResNet101\cite{resnet} & 75.39 & 89.14 & 82.74 & 51.87 & 69.28 & 77.65 & 82.42 & 88.08 & 90.89 & 86.26 & 85.13 & 60.60 & 66.30 & 74.05 & 67.76 & 58.77 \\ 
        $\text{S}^2$A-Net\cite{S2ANET} & ResNet101\cite{resnet} & 76.11 & 88.70 & 81.41 & 54.28 & 69.75 & 78.04 & 80.54 & 88.04 & 90.69 & 84.75 & 86.22 & 65.03 & 65.81 & 76.16 & 73.37 & 58.86\\ 
        ReDet\cite{han2021redet} & ReResNet\cite{han2021redet} & 76.25 & 88.79 & 82.64 & 53.97 & 74.00 & 78.13 & 84.06 & 88.04 & 90.89 & 87.78 & 85.75 & 61.76 & 60.39 & 75.96 & 68.07 & 63.59 \\ 
        \hline
        \multirow{3}{*}{\shortstack{Oriented\\ Rep\cite{li2022orientedreppoints}}} & ResNet50\cite{resnet} & 75.97 & 87.02 & 83.17 & 54.13 & 71.16 & 80.18 & 78.40 & 87.28 & \bf 90.90 & 85.97 & 86.25 & 59.90 & 70.49 & 73.53 & 72.27 & 58.97 \\ 
        & ResNet101\cite{resnet} & 76.52 & 89.53 & 84.07 & 59.86 & 71.76 & 79.95 & 80.03 & 87.33 & 90.84 & 87.54 & 85.23 & 59.15 & 66.37 & 75.23 & \bf 73.75 & 57.23 \\ 
        & Swin Tiny\cite{swin_2021_liu} & 77.63 & 89.11 & 82.32 & 56.71 & 74.95 & 80.70 & 83.73 & 87.67 & 90.81 & 87.11 & 85.85 & 63.60 & 68.60 & 75.95 & 73.54 & 63.76 \\ 
        \hline
        \multirow{7}{*}{\shortstack{Oriented\\ R-CNN\cite{orientedrcnn}}} & ResNet50\cite{resnet} & 75.87 & 89.46 & 82.12 & 54.78 & 70.86 & 78.93 & 83.00 & 88.20 & \bf 90.90 & 87.50 & 84.68 & 63.97 & 67.69 & 74.94 & 68.84 & 52.28 \\ 
        & ResNet101\cite{resnet} & 76.28 & 88.86 & 83.48 & 55.27 & 76.92 & 74.27 & 82.10 & 87.52 & \bf 90.90 & 85.56 & 85.33 & 65.51 & 66.82 & 74.36 & 70.15 & 57.28 \\
        & ARC-R50\cite{arcr50} & 77.35 & 89.40 & 82.48 & 55.33 & 73.88 & 79.37 & 84.05 & 88.06 & \bf 90.90 & 86.44 & 84.83 & 63.63 & 70.32 & 74.29 & 71.91 & 65.43 \\
        & Swin-T-RSP\cite{trsp} & 76.07 & 89.48 & 82.23 & 52.25 & 74.37 & 77.99 & 83.55 & 88.02 & \bf 90.90 & 87.42 & 84.92 & 57.70 & 65.99 & 73.88 & 69.30 & 63.07 \\
        & FasterNet T2\cite{FasterNetT2} & 76.17 & 89.51 & 78.20 & 52.65 & 73.11 & 78.36 & 83.39 & 87.80 & 90.86 & 87.58 & 84.60 & 62.22 & 61.89 & 75.04 & 71.03 & 66.28 \\
        & E-FormerV2 S2\cite{EFormerV2} & 76.70 & 89.55 & 84.12 & 53.39 & 74.40 & 80.70 & 84.84 & 87.92 & 90.89 & 87.44 & 84.47 & 60.88 & 67.43 & \bf 77.63 & 67.62 & 59.16 \\
        & DecoupleNet D2\cite{lu2024decouplenet} & 78.04 & 89.37 & 83.25 & 54.29 & 75.51 & 79.83 & 84.82 & \bf 88.49 & 90.89 & 87.19 & 86.23 & 66.07 & 65.53 & 77.23 & 72.34 & 69.62 \\ \hline
        \bf Ours & YOLOv8s &\bf 78.32 &\bf 90.51 & 73.48 &\bf 63.37 & 70.20 &\bf 82.90 &\bf 85.40 & 84.20 & 90.88 &\bf 88.80 & 70.29 &\bf 75.60 &\bf 87.90 & 72.13 & 69.30 & \bf 69.91 \\ \hline
    \end{tabular}
    \label{dota}
\end{table*}

Table \ref{dior} offers a focused comparison of detection methods on the DIOR-R\cite{dioraopg} dataset, emphasizing the balance between detection accuracy and computational efficiency. With a mAP of 67.32\%, our method outperforms the next best method by a margin of 0.24\%, while also presenting a reduced parameter count of 21.6M compared to DecoupleNet\cite{lu2024decouplenet}'s 23.3M. This efficiency in parameter usage, alongside lower FLOPs, highlights our model's suitability for applications where computational resources are at a premium.

\begin{table}[!ht]
    \centering
    \caption{Experimental results of the baseline network on the DIOR-R dataset. Complexities were tested by $800 \times 800$}
    \setlength{\tabcolsep}{2.5pt} 
    \begin{tabular}{cccc}
    \hline
        \bf Backbone & \bf Params.(M)$\downarrow$ & \bf FLOPs(G)$\downarrow$ & \bf mAP(\%)$\uparrow$ \\ \hline
        RetinaNet-O\cite{RotatedRetinanet} & 31.9  & 126.4  & 57.55  \\ 
        FR-O\cite{RotatedFasterrcnn} & 41.1  & 134.4  & 59.54  \\ 
        G.V.\cite{xu2020gliding} & 41.1  & 134.4  & 60.06  \\ 
        RoI Trans.\cite{ROITRANS} & 55.1  & 148.4  & 63.87  \\ 
        AOPG\cite{dioraopg} & - & - & 64.41  \\ 
        GGHL\cite{gghl} & - & - & 66.48  \\ 
        Oriented Rep\cite{li2022orientedreppoints} & 36.6  & 118.8  & 66.71  \\ 
        DCFL\cite{dcfl} & 36.1  & - & 66.80  \\ 
        0-RCNN-DecoupleNet D2\cite{lu2024decouplenet} & 23.3  & 92.3  & 67.08  \\ 
        \bf Ours &\bf 21.6  &\bf 52.8  &\bf 67.32 \\ \hline
    \end{tabular}
    \label{dior}
\end{table}

Table \ref{ucas} provides a targeted assessment of our model's performance on the UCAS-AOD dataset, which is pivotal for evaluating the detection of small targets within aerial imagery. Our method demonstrates a significant advancement in this domain, achieving a leading mAP of 97.86\%. This surpasses the performance of S2A-Net, which recorded a mAP of 89.99\%, and is particularly notable given the UCAS-AOD dataset's emphasis on small object detection.

\begin{table}[!ht]
    \centering
    \caption{Experimental results of the baseline network on the UCAS-AOD dataset}
    \begin{tabular}{cccc}
    \hline
        \bf Methods & \bf Car & \bf Airplane & \bf mAP50(\%)$\uparrow$ \\ \hline 
        RoI Trans.\cite{ROITRANS} & 88.02 & 90.02 & 89.02 \\ 
        SLA\cite{SLA} & 88.57 & 90.3 & 89.44 \\
        CFC-Net\cite{CFC-Net} & 89.29 & 88.69 & 89.49 \\
        TIOE-Det\cite{TIOE-Det} & 88.83 & 90.15 & 89.49 \\
        RIDet-O\cite{RIDet-Q} & 88.88 & 90.35 & 89.62 \\
        DAL\cite{DAL} & 89.25 & 90.49 & 89.87 \\
        $\text{S}^2\text{A}\text{-Net}$\cite{S2ANET} & 89.56 & 90.42 & 89.99 \\
        \bf Ours & \bf 96.32 & \bf 99.21 & \bf 97.86 \\ \hline
    \end{tabular}
    \label{ucas}
\end{table}

\begin{figure*}[htbp]
  \centering
  \includegraphics[width=1.0\textwidth]{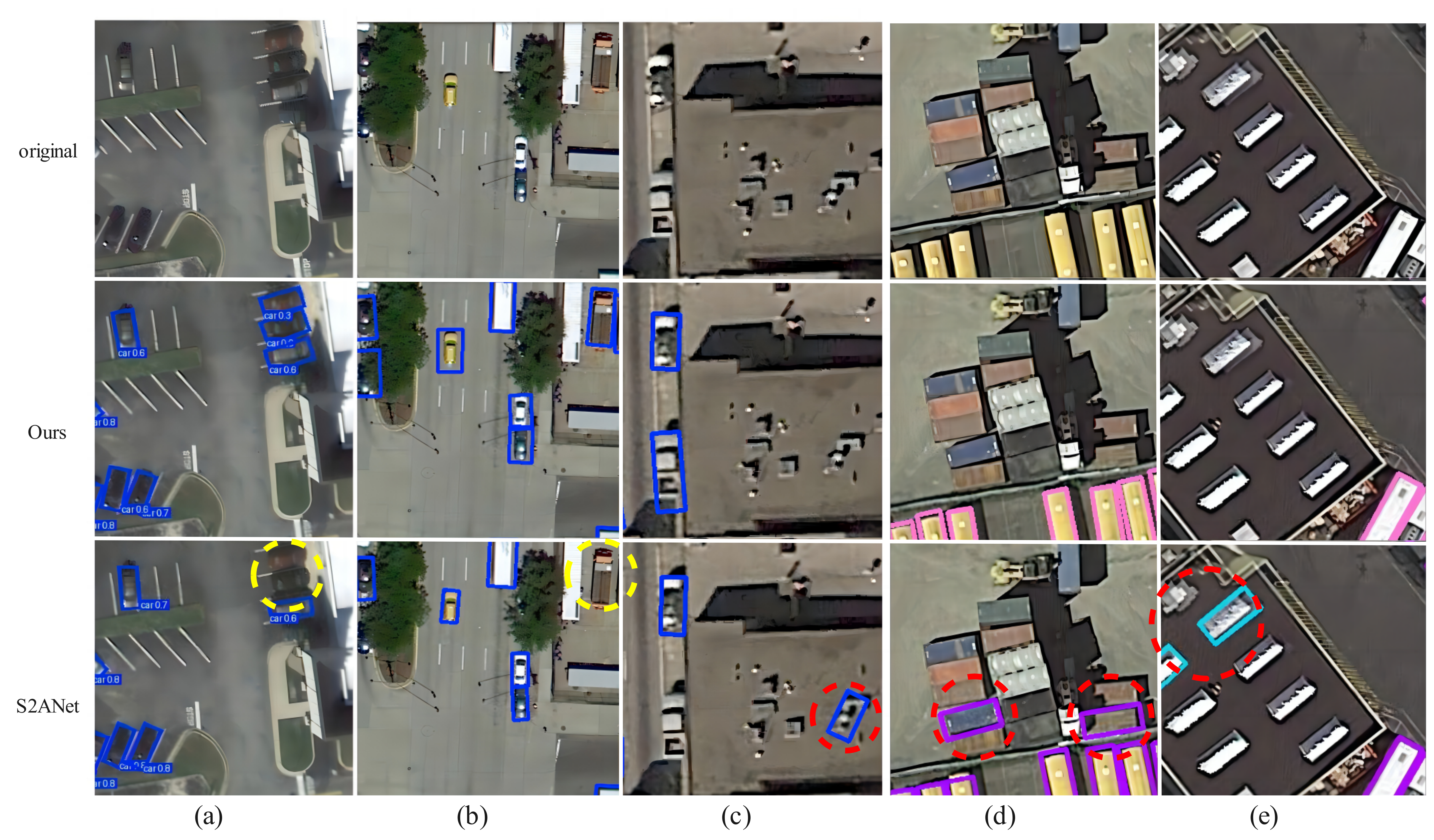}
  \caption{Comparative detection results displayed in three rows for clarity. The top row shows the original images from the UCAS-AOD (a, b, c) and DOTAv1.0 (d, e). The middle row corresponds to our method's predictions, while the bottom row represents $\text{S}^2\text{A}\text{-Net}$'s outcomes. Yellow dashed circles indicate missed detections, and red dashed circles highlight false detections.}
  \label{fig3}
\end{figure*}

Figure \ref{fig3} illustrates a comparative analysis between our model and $\text{S}^2\text{A}\text{-Net}$, highlighting the detection performance in scenarios with small objects and complex backgrounds. The results distinctly show that while $\text{S}^2\text{A}\text{-Net}$ exhibits several omissions and inaccuracies, our model consistently outperforms by accurately identifying and localizing targets. This demonstrates the enhanced precision and reliability of our approach, particularly effective in environments where small targets are embedded within intricate backgrounds.

Table \ref{parameter} presents a comparative analysis of various state-of-the-art methods for rotated bounding box object detection, focusing on the backbone architectures, parameter counts, and computational complexity measured in FLOPs. The table succinctly summarizes the key characteristics of each method, including the backbone used and the overall computational efficiency. This overview facilitates a direct comparison of the proposed method against existing approaches, highlighting its advantages in terms of reduced parameter count and lower computational requirements.

\begin{table}[!ht]
    \centering
    \caption{Overview of Methods, Backbone, and Parameters}
    \begin{tabular}{ccc|c}
    \hline
        \multirow{2}{*}{\bf Methods} & \multirow{2}{*}{\bf Backbone} & \multicolumn{2}{c}{\bf Overall /Backbone} \\ \cline{3-4}
        ~ & ~ & \bf Params.(M)$\downarrow$ & \bf FLOPs(G)$\downarrow$ \\ \hline
        RoI Trans.\cite{ROITRANS} & ResNet50\cite{resnet} & 55.1/23.3 & 225.3/86.1 \\ 
        G.V.\cite{xu2020gliding} & ResNet50\cite{resnet} & 41.1/23.3 & 211.3/86.1 \\ 
        \hline
        \multirow{3}{*}{\shortstack{Oriented\\ Rep\cite{li2022orientedreppoints}}} & ResNet50\cite{resnet} & 36.6/23.3 & 194.3/86.1 \\ 
        & ResNet101\cite{resnet} & 55.6/42.3 & 272.2/164 \\ 
        & Swin Tiny\cite{swin_2021_liu} & 37.3/27.5 & 200.5/95.4 \\ 
        \hline
        \multirow{7}{*}{\shortstack{Oriented\\ R-CNN\cite{orientedrcnn}}} & ResNet50\cite{resnet} & 41.1/23.3 & 211.4/86.1 \\ 
        & ResNet101\cite{resnet} & 60.1/42.3 & 289.3/164 \\ 
        & ARC-R50\cite{arcr50} & 74.4/- & 212.0/- \\ 
        & Swin-T-RSP\cite{trsp} & 44.8/27.5 & 215.7/195.4 \\ 
        & FasterNet T2\cite{FasterNetT2} & 30.0/12.7 & 160.3/40.0 \\ 
        & E-FormerV2 S2\cite{EFormerV2} & 29.2/12.0 & 145.1/26.8 \\ 
        & DecoupleNet D2\cite{lu2024decouplenet} & 23.3/6.2 & 142.4/23.1 \\ \hline
        \bf Ours & YOLOv8s & \bf 21.6/11.4 & \bf 93.0/76.3 \\ \hline
    \end{tabular}
    \label{parameter}
\end{table}

\begin{figure}[htbp]
  \centering
  \includegraphics[width=0.48\textwidth]{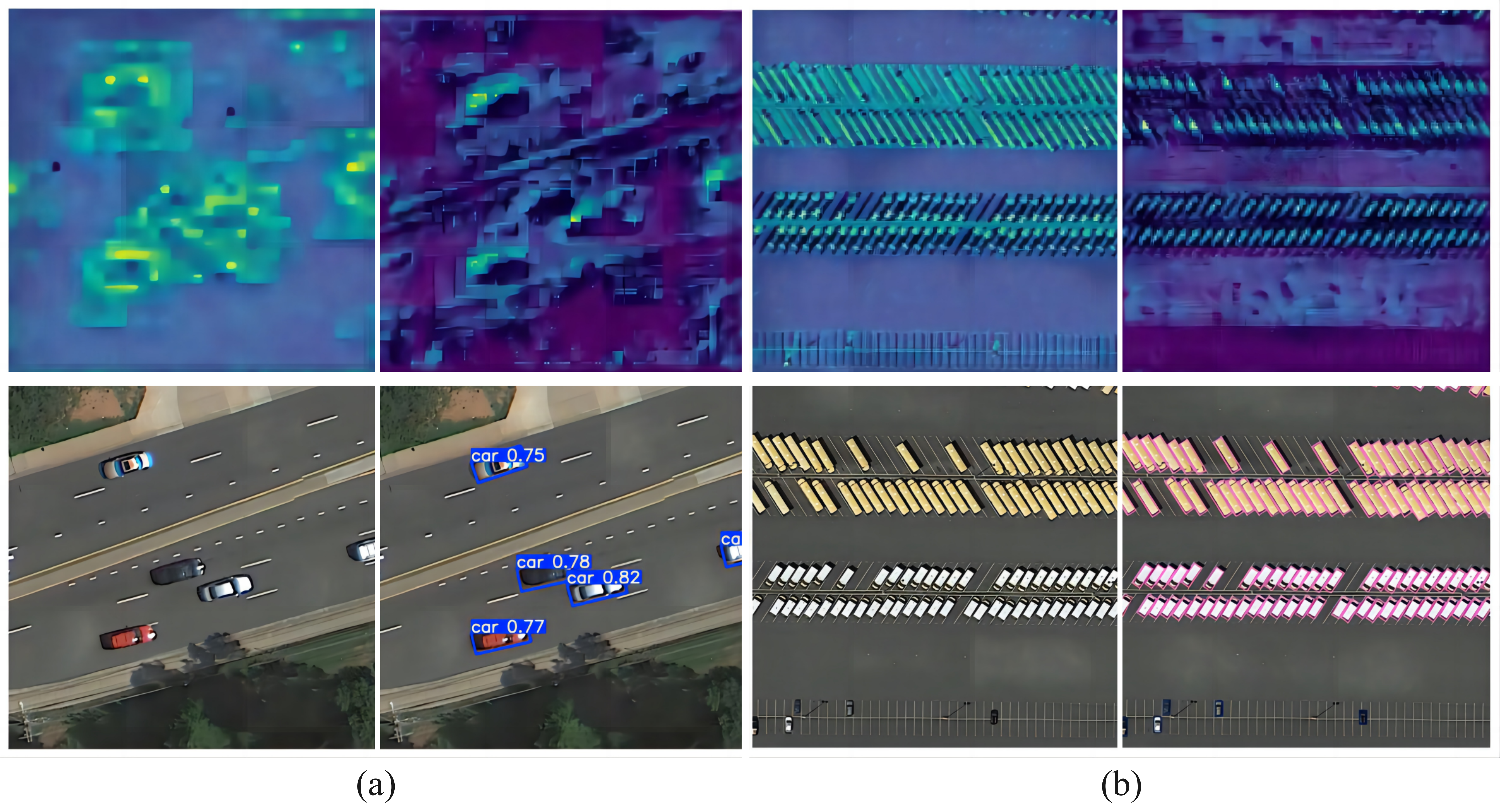}
  \caption{ The top row displays feature maps from the C2f-WTC (left) and ASFP (right) modules.The bottom row presents the original images (left) alongside the corresponding prediction results (right)for (a) UCAS-AOD and (b) DOTAv1.0.}
  \label{fig4}
\end{figure}

On the UCAS-AOD\cite{ucas} dataset and DOTAv1.0\cite{xia2018dota}, we confirmed the efficacy of our proposed modules by analyzing the output feature maps, as depicted in Figure \ref{fig4}. The clear localization of small objects post-processing substantiates the viability of our approach.

\subsection{Ablation Studies}
To validate the effectiveness of each component in our method, we conducted an ablation study on the UCAS-AOD dataset\cite{ucas}. Here, \textbf{S} denotes the ASFP module, \textbf{W} denotes the replacement of the backbone's original C2f structure with the C2f-WTC module, while \textbf{G} signifies the substitution of the neck's C2f structure with the C2f-GDC module.

\begin{table}[!ht]
\centering
\caption{Comparison of detection performance and model complexity.}
\begin{tabular}{cccccc}
\hline
\textbf{Backbone} & \textbf{S} & \textbf{W}& \textbf{G}& \textbf{mAP50(\%)} & \textbf{Params(M)} \\ \hline
\multirow{7}{*}{YOLOv8s} & & & & 95.83 & \bf 11.4 \\ 
 & $\checkmark$ & & & 97.03 & 18.89 \\
 &  & $\checkmark$ & & 96.94 & 16.67 \\
 & $\checkmark$ &$\checkmark$ & & 97.84 & 23.29 \\
 & $\checkmark$ & &$\checkmark$ & 96.65 & 17.87 \\
 & &$\checkmark$ &$\checkmark$ & 96.49 & 15.20 \\
 & $\checkmark$ &$\checkmark$ &$\checkmark$ & \bf 97.86 & 21.63 \\ \hline
\end{tabular}
\label{table3}
\end{table}

As shown in Table \ref{table3}, introducing the ASFP module increased the mAP by 0.52\%, with only a slight increase in model parameters and FLOPs. The C2f-WTC module significantly enhanced multi-scale feature extraction capability, achieving an accuracy of 97.84\%. Finally, the C2f-GDC module improved accuracy while maintaining a low parameter count. When combining all modules, the proposed method demonstrated excellent performance in both accuracy and real-time processing, verifying the effectiveness and synergy of each module.
\section{CONCLUSION}
In this study, we proposed a YOLOv8-based rotated bounding box object detection method that enhances small object detection capabilities while maintaining a low-parameter architecture. By integrating the ASFP, C2f-WTC, and C2f-GDC modules, the proposed method achieves significant improvements in detection accuracy and efficiency. The experimental results demonstrate the effectiveness of the proposed method, providing strong support for practical applications in resource-constrained environments.
\bibliographystyle{unsrt}
\bibliography{my}

\end{document}